\documentclass[11pt]{article}

\usepackage[margin=1in]{geometry}
\usepackage{authblk}
\usepackage{fancyhdr}
\usepackage{setspace}
\usepackage{hyperref}
\usepackage{lipsum}
\usepackage{graphicx} 
\usepackage{etoolbox} 

\fancypagestyle{firstpage}{
  \fancyhf{}
  
  \fancyhead[L]{%
  \begin{minipage}{\textwidth}
  \footnotesize
  \textbf{This is the accepted version of the following paper:} \\
  \copyright~2025 IEEE. Accepted to the 2025 IEEE International Conference on Content-Based Multimedia Indexing (CBMI). \\
  Personal use of this material is permitted. Permission from IEEE must be obtained for all other uses, including reprinting/republishing this material for advertising or promotional purposes, creating new collective works, for resale or redistribution to servers or lists, or reuse of any copyrighted component of this work.
  \end{minipage}
}
}
\title{Beyond CNNs: Efficient Fine-Tuning of Multi-Modal LLMs for Object Detection on Low-Data Regimes}
\author[1]{Nirmal Elamon}
\author[2]{Rouzbeh Davoudi}
\affil[1]{Artificial Creative intelligence (ACI), Expedia Group, nirmalelamon13@gmail.com}
\affil[2]{Artificial Creative intelligence (ACI), Expedia Group, rouzbeh.davoudi@gmail.com}

\begin{document}
\onehalfspacing  
\maketitle

\thispagestyle{firstpage}

\begin{abstract}
The field of object detection and understanding is rapidly evolving, driven by advances in both traditional CNN-based models and emerging multi-modal large language models (LLMs). While CNNs like ResNet and YOLO remain highly effective for image-based tasks, recent transformer-based LLMs introduce new capabilities such as dynamic context reasoning, language-guided prompts, and holistic scene understanding. However, when used out-of-the-box, the full potential of LLMs remains underexploited, often resulting in suboptimal performance on specialized visual tasks. 

In this work, we conduct a comprehensive comparison of fine-tuned traditional CNNs, zero-shot pre-trained multi-modal LLMs, and fine-tuned multi-modal LLMs on the challenging task of artificial text overlay detection in images. A key contribution of our study is demonstrating that LLMs can be effectively fine-tuned on \textbf{very limited data (fewer than 1,000 images)} to achieve up to \textbf{36\% accuracy improvement}, matching or surpassing CNN-based baselines that typically require orders of magnitude more data. 

By exploring how language-guided models can be adapted for precise visual understanding with minimal supervision, our work contributes to the broader effort of \textbf{bridging vision and language}, offering novel insights into efficient cross-modal learning strategies. These findings highlight the adaptability and data efficiency of LLM-based approaches for real-world object detection tasks and provide actionable guidance for applying multi-modal transformers in low-resource visual environments. To support continued progress in this area, we have made the code used to fine-tune the models available in our GitHub, enabling future improvements and reuse in related applications.
\end{abstract}

\vspace{0.5em}
\noindent
\footnotesize
\textbf{https://github.com/NirmalElamon/Phi3.5-vision-finetuning} \\
\textbf{Copyright Notice:} \copyright~2025 IEEE. Personal use of this material is permitted. Accepted to CBMI 2025. Permission required for all other uses.
\normalsize
\vspace{1em}



\section{Introduction}
Object detection has always been a cornerstone of computer vision, enabling applications and use cases ranging across autonomous driving ,  healthcare, medical imaging, retail, e-commerce and content moderation. Traditionally, convolutional neural network (CNN) based architectures such as ResNet, Faster R-CNN, and YOLO have dominated this field, powering robust image-based feature extraction and pattern recognition capabilities. However, the latest development of transformer-based models—especially multi-modal large language models (LLMs)—has introduced new paradigms for understanding and aligning the visual and language data with better context. 

These multi-modal LLMs, trained on vast amount of paired image-text data, offer intriguing capabilities beyond traditional vision pipelines. Their ability to reason over complex scenes, interpret textual cues, and adapt to diverse and specific downstream tasks with minimal supervision opens up new avenues in cross-modal learning. Yet, despite their capabilities, these models are often used as off-the-shelf models through zero-shot or few-shot settings without targeted fine-tuning, leading to under-performing results on domain-specific visual tasks. 

One such task that poses a unique challenge is the detection of artificial text overlays in images—a problem relevant to digital media verification, content moderation, and document analysis. This task requires careful and nuanced scene understanding and fine-grained discrimination between natural and synthetic textual content, making it an ideal benchmark for evaluating the adaptability of vision-language models. Additionally, It also requires models to accurately interpret contextual cues, such as the relationship between the text and the object it appears on. For instance, without proper fine-tuning, a pre-trained LLM might mistakenly flag the word "NETFLIX" displayed on a television screen as artificial text, despite it being a natural part of the scene. 

In this paper, we investigate the effectiveness of multi-modal LLMs through different strategies and compare them with the traditional CNN approach for artificial text overlay detection. We conduct a systematic comparison across three modeling paradigms: (1) traditional CNNs fine-tuned on task-specific data, (2) zero-shot pre-trained LLMs using two different prompt engineering strategies, and (3) fine-tuned LLMs trained with a small number of annotated images. Our experiments show that fine-tuned LLMs not only surpass the performance of both zero-shot LLMs and traditional CNN-based approaches, but also achieve this with remarkably few training samples (less than 1000 images)—requiring minimal human annotation for effective fine-tuning. 

\section{Related Work}
The field of object detection has seen substantial advancements in the past decade, mainly driven by the success of convolutional neural networks (CNN) and, more recently, the introduction of transformer-based architectures. Traditional CNN-based object detection methods can be broadly classified into one-stage detectors, such as \cite{yolo} (You Only Look Once) and \cite{SSD} (Single Shot MultiBox Detector), which prioritize speed and efficiency, and two-stage detectors, such as R-CNN and Faster R-CNN \cite{FastRCNN}, which typically offer higher accuracy through a region proposal and refinement process. 

When it comes to the specific task of detecting artificial text overlay, conventional CNN approaches face notable limitations. Most prior work in this domain focuses on general scene text detection, identifying the presence of any text within an image, rather than explicitly distinguishing between naturally embedded and artificially overlaid text. For example, studies such as \cite{Rizky} and \cite{Mansoor} explore the use of pre-trained or fine-tuned CNN models for detecting textual regions in natural images. However, these methods do not address the more nuanced challenge of detecting synthetically overlaid text, which requires a deeper contextual understanding of the image, the semantics of the scene, and the spatial relationship between the text and its background. As a result, traditional CNN-based models often fail to accurately distinguish artificial overlays from naturally occurring text, especially in visually complex or ambiguous scenarios. 

To overcome the limitations of traditional CNN-based methods, recent research has shifted toward leveraging large language models (LLMs) in multi-modal settings, where visual and textual modalities are processed jointly. Models such as CLIP \cite{Radford}, Phi3 \cite{Phi3}, and GPT-4V \cite{down_llm} exemplify this new class of vision language architectures, which are pre-trained on large-scale image-text pairs to align visual features with semantic language representations. These models exhibit strong capabilities in reasoning about visual content using natural language, making them particularly well-suited for tasks that require high-level semantic understanding and contextual awareness, such as text detection in images. 

Unlike CNNs that primarily operate on low-level visual features, multimodal LLMs can incorporate language-guided prompts and scene-level context to more accurately detect and interpret text within an image. This results in a more flexible and robust understanding of complex visual scenarios, especially in cases where isolated visual cues are ambiguous or insufficient. For example, recent work such as \cite{Lumos} demonstrates how LLMs can be employed to detect text in diverse visual environments. However, while these approaches have advanced the field of general text detection, they primarily focus on identifying any textual content present in an image and do not specifically address the more challenging problem of detecting artificially overlaid text, where distinguishing between natural and synthetic text requires deeper contextual reasoning and spatial understanding. 

In light of these limitations, our work seeks to bridge the gap by demonstrating that artificial text overlay detection can be effectively tackled through a data-efficient fine-tuning strategy leveraging multimodal LLMs. Our experiments reveal that off-the-shelf LLMs and fine-tuned CNNs often suffer from high false positive rates, frequently misclassifying naturally embedded text as artificially overlaid. In contrast, we show that with targeted fine-tuning—performed over just a few epochs and using fewer than 1,000 annotated samples—multi-modal LLMs can be adapted to accurately distinguish between natural and synthetic text. This approach not only reduces misclassification but also consistently outperforms traditional CNN-based methods, highlighting the effectiveness of language-guided models in handling complex visual-textual tasks with minimal supervision.

\section{Methodology}
We evaluate and compare the following model variants on a shared, curated testing dataset specifically designed for the task of detecting artificial text overlays in images.

\subsection{Fine-tuned CNN}
\begin{figure}[htbp]
\centerline{\includegraphics[width=0.41\linewidth]{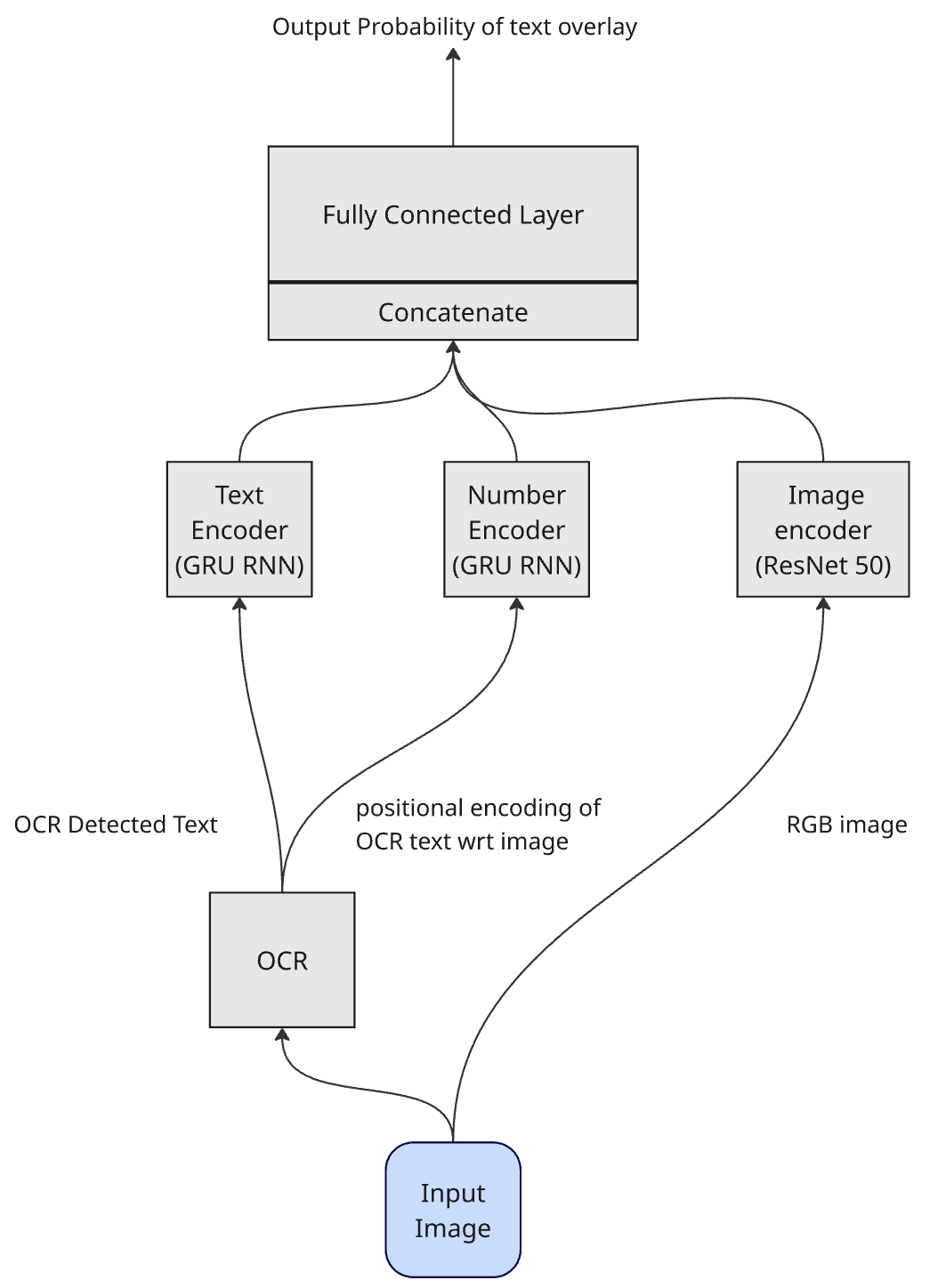}}
\caption{Traditional fine-tuned CNN model}
\label{fig:CNN}
\end{figure}

We employ a conventional CNN architecture (OCR + ResNet \cite{Kaiming}), fine-tuned on a large labeled dataset of 10,000 artificial text overlay images (Figure \ref{fig:CNN}).  The architecture integrates both visual and textual cues to make informed binary classification decisions (text overlay vs. no overlay) by leveraging OCR outputs, positional metadata, and deep image features. The pipeline begins with the input image being passed through a ResNet-based OCR module. Each detected text element is also associated with a positional encoding vector, representing its location (e.g., x/y coordinates, width, height) and relative layout on the image canvas.  This encoding captures: Vertical and horizontal alignment patterns, Distribution density (e.g., tightly packed overlay text vs. sparse natural signage) and Hierarchical grouping (e.g., header/footer text blocks). These positional cues are critical, as overlaid text often exhibits uniform alignment and fixed margins, unlike organic scene text. The extracted OCR text and the associated positional encodings are processed through two separate GRU (Gated Recurrent Unit) based encoders. Both encoders allow the model to learn contextual dependencies and recurrent structures in the text and layout — useful in differentiating patterned artificial overlays from incidental scene text.

 In parallel, image features are generated by passing the input image through an Image Encoder based on ResNet 50 model. The resulting OCR text feature, positional encoded number feature and the image feature are all concatenated together and send into the final fully connected layer which outputs the probability of text overlay in the given image. The entire architecture is trained end-to-end on a curated dataset of 10,000 labeled images, evenly distributed across artificial text overlays, natural text, and no-text conditions. The model is optimized using a binary cross-entropy loss, which is suitable for this binary classification task

\subsection{Pretrained Zero-Shot LLM}
\begin{figure}[htbp]
\centerline{\includegraphics[width=0.48\linewidth]{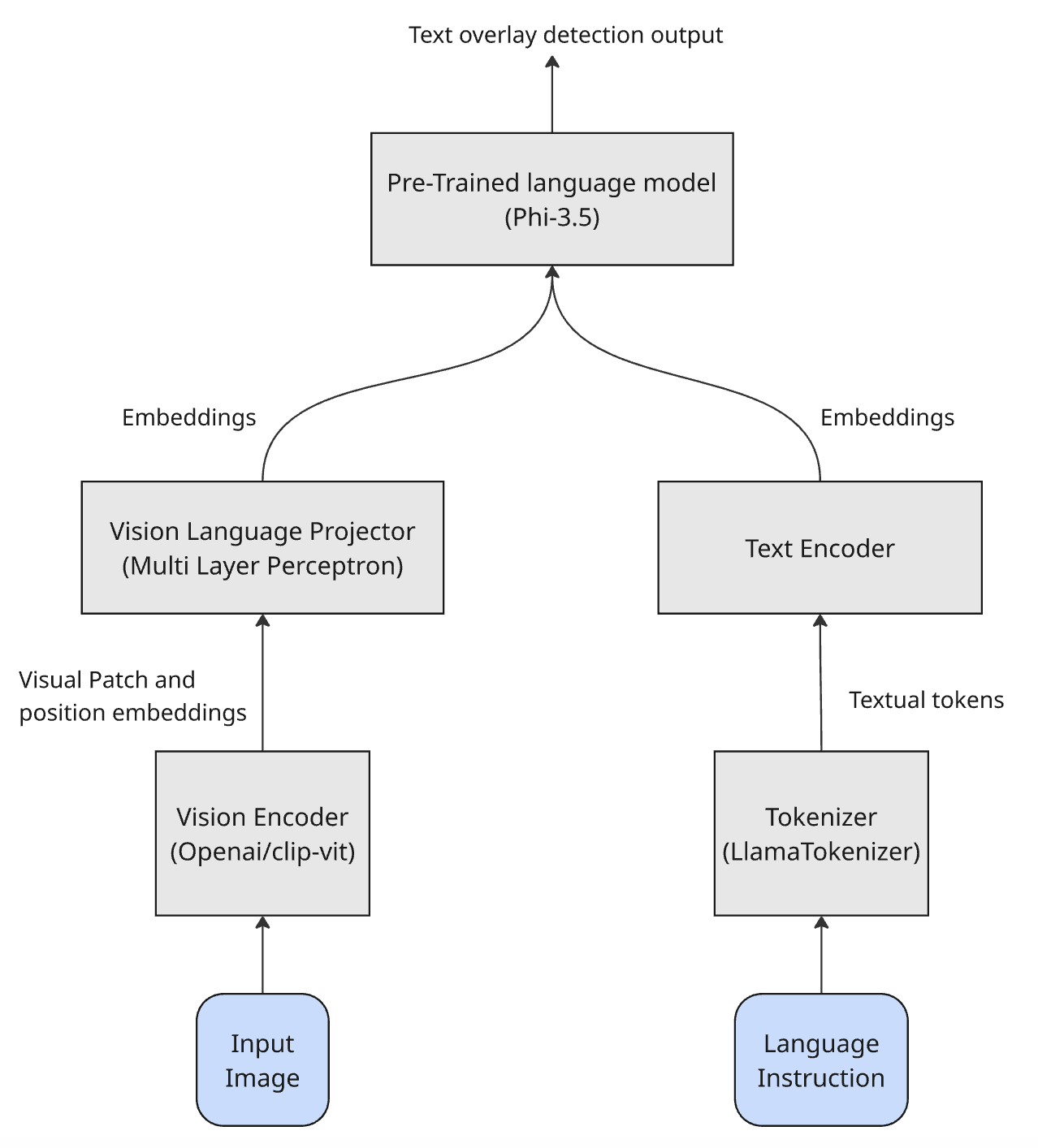}}
\caption{Pre-trained Phi-3.5 model}
\label{fig:pre_tuned_phi_3}
\end{figure}

 A pre-trained multi-modal LLM (Phi3.5-vision) \cite{Phi3} is used without any task-specific fine-tuning. The model leverages its existing visual and language understanding to perform object detection using a single descriptive prompt, relying on zero-shot generalization. The structure of this model is shown in figure  \ref{fig:pre_tuned_phi_3}. We utilize the instruction-tuned, off-the-shelf Phi-3.5 Vision Instruct model for artificial text overlay detection. The input image is first processed by a vision encoder based on Openai Clip vit that extracts visual features enriched with positional embeddings. Concurrently, a task-specific prompt is crafted to guide the model in identifying artificial text overlays and their associated cues; this prompt is then tokenized. The visual features are passed through a multilayer perceptron (MLP), while the tokenized prompt is processed by a text encoder to produce corresponding embeddings. These image and text embeddings are subsequently fed into the pre-trained Phi-3.5 language model, which outputs the final text overlay detection prediction.

\subsection{Pretrained Zero-Shot LLM with sequential prompting}
\begin{figure}[htbp]
\centerline{\includegraphics[width=0.48\linewidth]{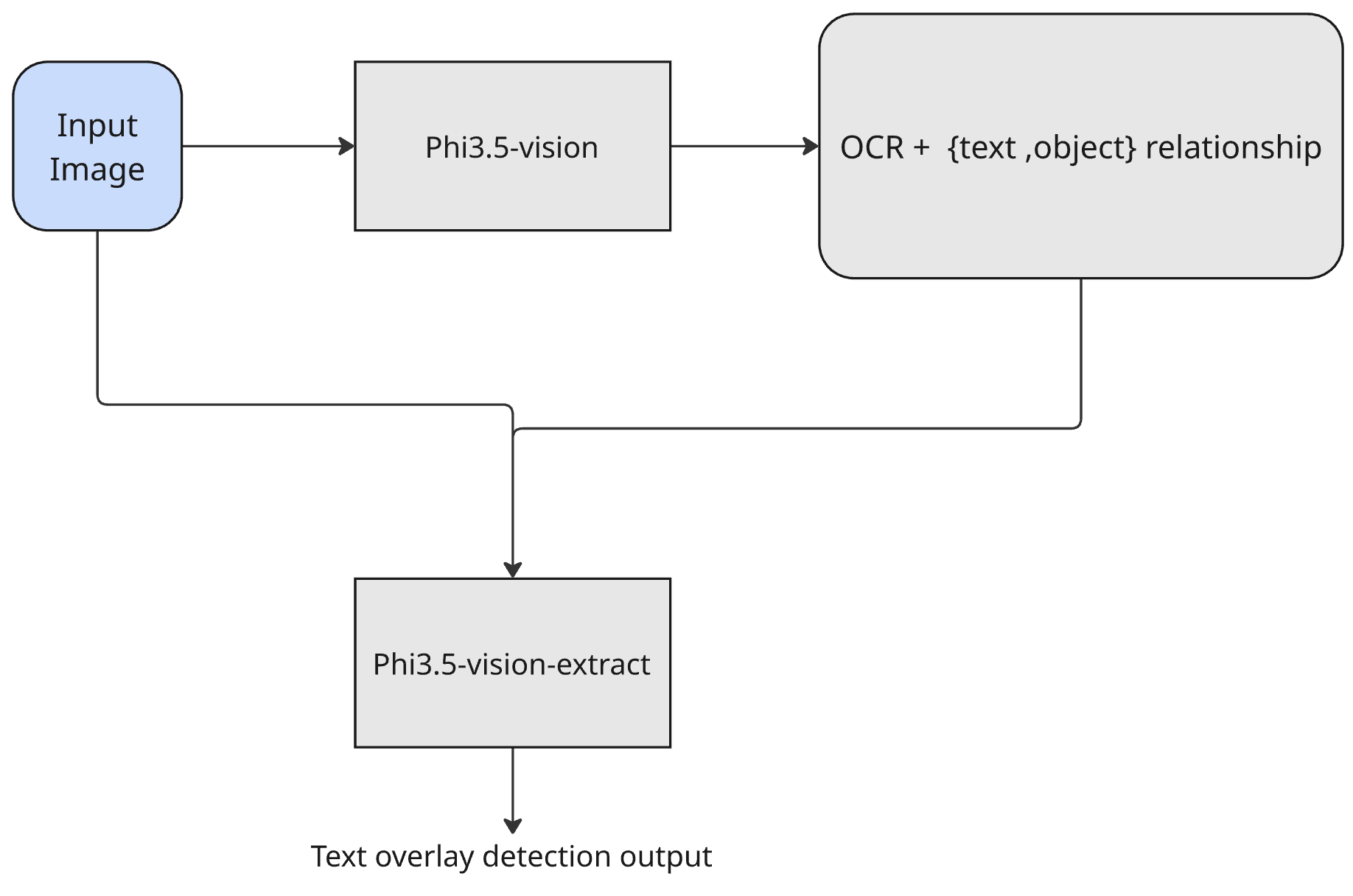}}
\caption{Phi-3.5 model with sequential prompting}
\label{fig:seq_phi_3}
\end{figure}

This approach extends the baseline zero-shot LLM setup (using Phi3.5-Vision) \cite{Phi3} by introducing a sequence of context-aware prompts to iteratively refine the detection output as shown in Figure \ref{fig:seq_phi_3}. Rather than updating model weights, it improves performance through prompt chaining. Initially, Phi3.5-Vision \cite{Phi3} is used to perform optical character recognition (OCR), extracting all visible text in the image along with contextual information about the objects on which the text appears. This intermediate response, along with the original image, is then passed back to the same model to evaluate which of the detected texts are artificially overlaid. By incorporating a second round of reasoning enriched with object-text relational context, this sequential prompting strategy enhances the model’s ability to differentiate between naturally embedded and artificially added text. 

The first zero-shot prompting inference is broken down into two smaller tasks:
\begin{enumerate}
    \item Identify all text and objects present in the image. The set of detected objects in the image $c_i$ is denoted as $O_i = \{o_1, o_2, \ldots, o_{n_i}\}$, where $n_i$ is the number of objects in frame $c_i$. The set of text entities in the image $c_i$ is denoted as $T_i = \{t_1, t_2, \ldots, t_{m_i}\}$, where $m_i$ is the number of text entities in frame $c_i$.
    \item Once the text entities and objects are detected, the model again determines the spatial relationships between the detected objects and the text entities in each frame $c_i$. Let $R_i = \{r_1, r_2, \ldots, r_{n_{r_i}}\}$ be the set of relationships identified in frame $c_i$, where $n_{r_i}$ is the number of relationships in frame $c_i$.

\end{enumerate}

Once the $O_i$, $T_i$, and $R_i$ are identified, they are provided as input along with the image itself for the second inference stage, which aims to detect artificial text overlays. This ensures that the model is well-informed during the second inference about all textual content and its spatial relationships with background objects, thereby enabling a more accurate determination of text overlays.

\subsection{Fine-tuned LLM}

The multi-modal LLM, Phi-3.5 Vision \cite{Phi3}, is fine-tuned on a domain-specific dataset consisting of 1,000 annotated images containing artificial text overlays. The annotations indicate whether a given image contains overlaid synthetic text, enabling supervised binary classification. Fine-tuning is performed for 2 epochs using binary cross-entropy loss, with early stopping based on validation accuracy to prevent overfitting.

Despite the relatively small size of the dataset, the lightweight nature of Phi-3.5 Vision, combined with its instruction-tuned architecture, enables efficient specialization toward the task of artificial text overlay detection. This adaptation leads to a marked improvement in detection accuracy compared to the zero-shot performance of the base model. The enhanced performance demonstrates the model's ability to generalize the learned cues of artificial text—such as unnatural alignment, occlusion patterns, and inconsistent font rendering—when provided with minimal supervision.

\begin{figure}[htbp]
\centerline{\includegraphics[width=0.78\linewidth]{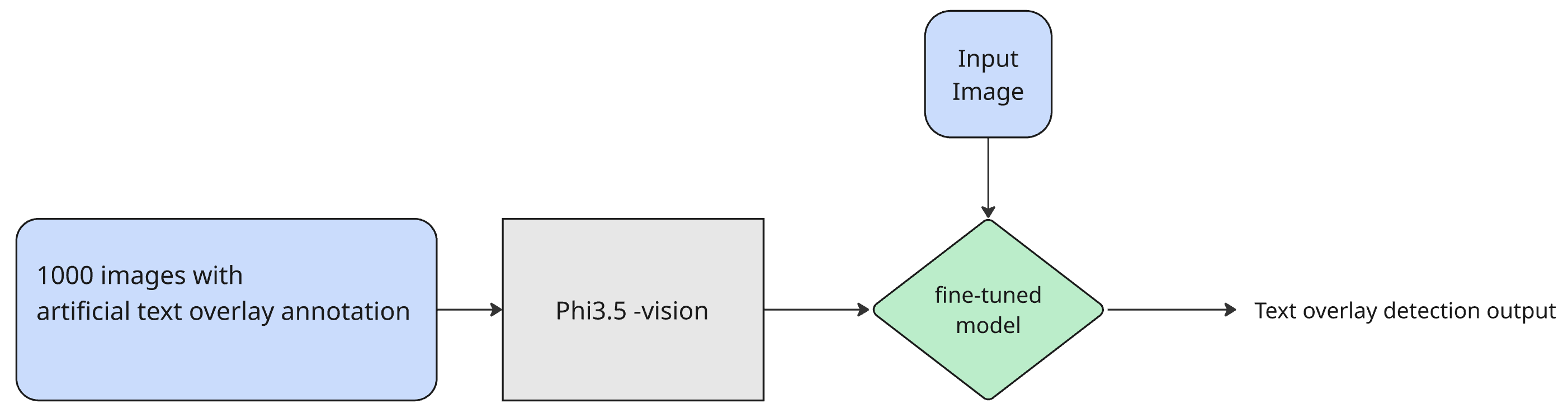}}
\caption{Fine-tuned Phi-3.5 model}
\label{fig:fine_tuned_phi_3}
\end{figure}

The complete fine-tuning pipeline is illustrated in Figure~\ref{fig:fine_tuned_phi_3}. This process highlights the practicality of tailoring a general-purpose vision-language model to a highly specific visual reasoning task with minimal annotation effort.

\section{Dateset for Finetuning and Evaluation}
For the fine-tuning process, we curated a high-quality dataset comprising 1,000 images, evenly distributed across three distinct categories: (1) images with artificial text overlays, (2) images containing naturally embedded text (e.g., street signs, book covers, product packaging), and (3) images without any visible text. This careful class balancing ensures that the model is exposed to a diverse range of visual-textual scenarios and prevents bias toward any particular category during training.

Each image in the dataset was manually annotated by human reviewers to ensure the accuracy and consistency of ground truth labels. Annotators were provided with clear guidelines to distinguish between artificially overlaid text—typically characterized by inconsistent font styling, unnatural alignment, or visual detachment from the background—and naturally embedded text that is physically integrated into the scene. Images with no text served as a necessary control class to discourage false positives during inference.

In addition to the training set, a separate evaluation dataset was constructed, also consisting of 1,000 images with the same balanced distribution across the three categories. This mirrored composition enables a fair and comprehensive assessment of model performance across all text visibility scenarios. By maintaining this balance in both training and evaluation sets, the model’s accuracy metrics reflect its true generalization ability and not overfitting to a specific text type or visual pattern.

\section{Hyperparameters and Training Setup}
To fine-tune the Phi-3.5 Vision model for the text overlay detection task, we adopted a lightweight but carefully designed training configuration optimized for stability and efficient use of limited data. The model was trained using DeepSpeed for distributed optimization and tensor memory efficiency, with training monitored through TensorBoard logs.

During fine-tuning, the vision tower was frozen to retain general-purpose image representation capabilities, while the language model weights were kept trainable to allow task-specific adaptation. To balance the contributions of textual and visual streams, the image projector layers were fine-tuned, ensuring improved alignment between visual embeddings and text tokens.

Key hyperparameters were chosen to ensure stability with small batch sizes and high-dimensional vision-language inputs. We used a per-device batch size of 1 with gradient accumulation of 2, resulting in an effective batch size of 2. Training was conducted for 2 epochs with a cosine learning rate schedule, a learning rate of 2e-4, and warmup ratio of 0.03. Regularization was introduced via weight decay (0.0) and gradient checkpointing to reduce memory overhead. Mixed precision was enabled with bfloat16 (bf16) while explicitly disabling fp16 to avoid numerical instabilities observed in preliminary runs.

For image handling, we applied 16 random crops per sample to expose the model to varied spatial contexts. Additional runtime optimizations included enabling TF32 matrix multiplications, setting 2 dataloader workers, and disabling FlashAttention v2 due to compatibility concerns. Logging was performed at every step to facilitate granular tracking of convergence.

Overall, this setup ensured that fine-tuning could be performed efficiently on a modestly sized dataset while maintaining stability, generalization, and reproducibility across runs.

\begin{table}[htbp]
\caption{Hyperparameters and Training Configuration for Fine-tuned Phi-3.5 Vision}
\begin{center}
\begin{tabular}{|l|l|}
\hline
\textbf{Parameter} & \textbf{Value} \\
\hline
Training epochs & 2 \\
\hline
Per-device batch size & 1 \\
\hline
Gradient accumulation steps & 2 \\
\hline
Effective batch size & 2 \\
\hline
Learning rate & 2e-4 \\
\hline
Scheduler & Cosine decay with 0.03 warmup \\
\hline
Weight decay & 0.0 \\
\hline
Precision & bf16 enabled, fp16 disabled \\
\hline
Vision tower & Frozen \\
\hline
LLM weights & Trainable \\
\hline
Image projector & Fine-tuned \\
\hline
Number of crops per image & 16 \\
\hline
Gradient checkpointing & Enabled \\
\hline
TF32 & Enabled \\
\hline
FlashAttention v2 & Disabled \\
\hline
Logging & Every step (TensorBoard) \\
\hline
\end{tabular}
\label{tab:hyperparams}
\end{center}
\end{table}

\section{Model Comparison}
The results presented in Table~\ref{tab1} clearly demonstrate the superior performance of the fine-tuned LLM across all three evaluation metrics: precision, recall, and accuracy. Achieving a precision of 0.98, recall of 0.84, and accuracy of 0.83, the fine-tuned Phi-3.5 Vision model consistently outperforms all other baselines. This highlights the effectiveness of task-specific adaptation even with a relatively small dataset of just 1,000 annotated samples. The fine-tuning enables the model to learn subtle visual and spatial cues characteristic of artificial text overlays, such as uniform text alignment, occlusion artifacts, and unnatural positioning relative to background elements.

In contrast, the pre-trained Phi-3.5 Vision model without fine-tuning, when evaluated in a zero-shot setting, performs notably worse, with an accuracy of 0.60. Although it demonstrates a relatively high recall of 0.80—suggesting it is able to detect many instances of overlaid text—it suffers from poor precision (0.66), indicating a tendency to misclassify natural or no-text images as containing overlays. This behavior is expected, as the model lacks explicit exposure to the task-specific signal and therefore struggles to distinguish between visually similar patterns.

The sequential prompting variant of the pre-trained model shows modest improvements in precision (0.75) due to the inclusion of intermediate reasoning steps that incorporate contextual relationships between objects and text. However, the recall (0.51) and overall accuracy (0.54) remain low, demonstrating the limitations of relying solely on prompt chaining without parameter updates. While sequential prompting introduces valuable relational context (e.g., text-object positioning), the lack of weight adaptation limits its ability to consistently generalize across varied text overlay styles.

The traditional CNN model, despite being trained on a substantially larger dataset of 10,000 annotated images, performs the worst in terms of both precision (0.54) and accuracy (0.47). Although it achieves a reasonable recall of 0.78—likely due to its aggressive positive classification threshold—it fails to effectively separate artificial overlays from natural text. This can be attributed to its limited capacity to reason over complex semantic relationships and compositional cues in visual scenes. Moreover, the CNN model's dependence on engineered features such as OCR outputs and positional encodings constrains its adaptability when faced with ambiguous or noisy input data.

These comparative results collectively highlight the power of fine-tuning lightweight, instruction-tuned multi-modal LLMs for specialized visual understanding tasks. Unlike traditional CNNs or prompt-only inference strategies, fine-tuned LLMs are able to jointly reason over spatial, semantic, and textual signals in a unified manner—offering high performance with minimal supervision.

\begin{table}[htbp]
\caption{Model Comparison and Results}
\begin{center}
\begin{tabular}{|c|c|c|c|}
\hline
\textbf{Model} & \textbf{\textit{Precision}} & \textbf{\textit{Recall}} & \textbf{\textit{Accuracy}} \\
\hline
Fine-tuned LLM & 0.98 & 0.84 & 0.83 \\
\hline
Pre-trained LLM, seq re-prompting & 0.75 & 0.51 & 0.54 \\
\hline
Pre-trained LLM & 0.66 & 0.80 & 0.60 \\
\hline
Traditional CNN model & 0.54 & 0.78 & 0.47 \\
\hline
\end{tabular}
\label{tab1}
\end{center}
\end{table}

In addition to the baselines reported above, we also evaluated a widely adopted off-the-shelf multi-modal LLMs—Qwen2.5-VL-7B-Instruct under the same experimental setup. Both models show competitive performance compared to the zero-shot and sequential prompting baselines, though they still fall short of the fine-tuned Phi-3.5 Vision. Specifically, Qwen2.5-VL-7B-Instruct demonstrates balanced precision and recall, achieving an accuracy of 0.78. These results confirm that off-the-shelf instruction-tuned vision-language models can provide a strong baseline without task-specific adaptation. However, they still struggle with domain-specific subtleties—such as differentiating natural text from artificial overlays—that the fine-tuned Phi-3.5 Vision model captures more reliably. This underscores the importance of lightweight fine-tuning even when starting from powerful instruction-aligned models.

\begin{table}[htbp]
\caption{Comparison of Fine-tuned LLM with an Off-the-Shelf Pretrained LLM}
\begin{center}
\begin{tabular}{|c|c|c|c|}
\hline
\textbf{Model} & \textbf{\textit{Precision}} & \textbf{\textit{Recall}} & \textbf{\textit{Accuracy}} \\
\hline
Fine-tuned LLM (Phi-3.5 Vision) & 0.98 & 0.84 & 0.83 \\
\hline
Qwen2.5-VL-7B-Instruct & 0.82 & 0.75 & 0.78 \\
\hline
\end{tabular}
\label{tab2}
\end{center}
\end{table}

\begin{figure}[htbp]
\centerline{\includegraphics[width=0.78\linewidth]{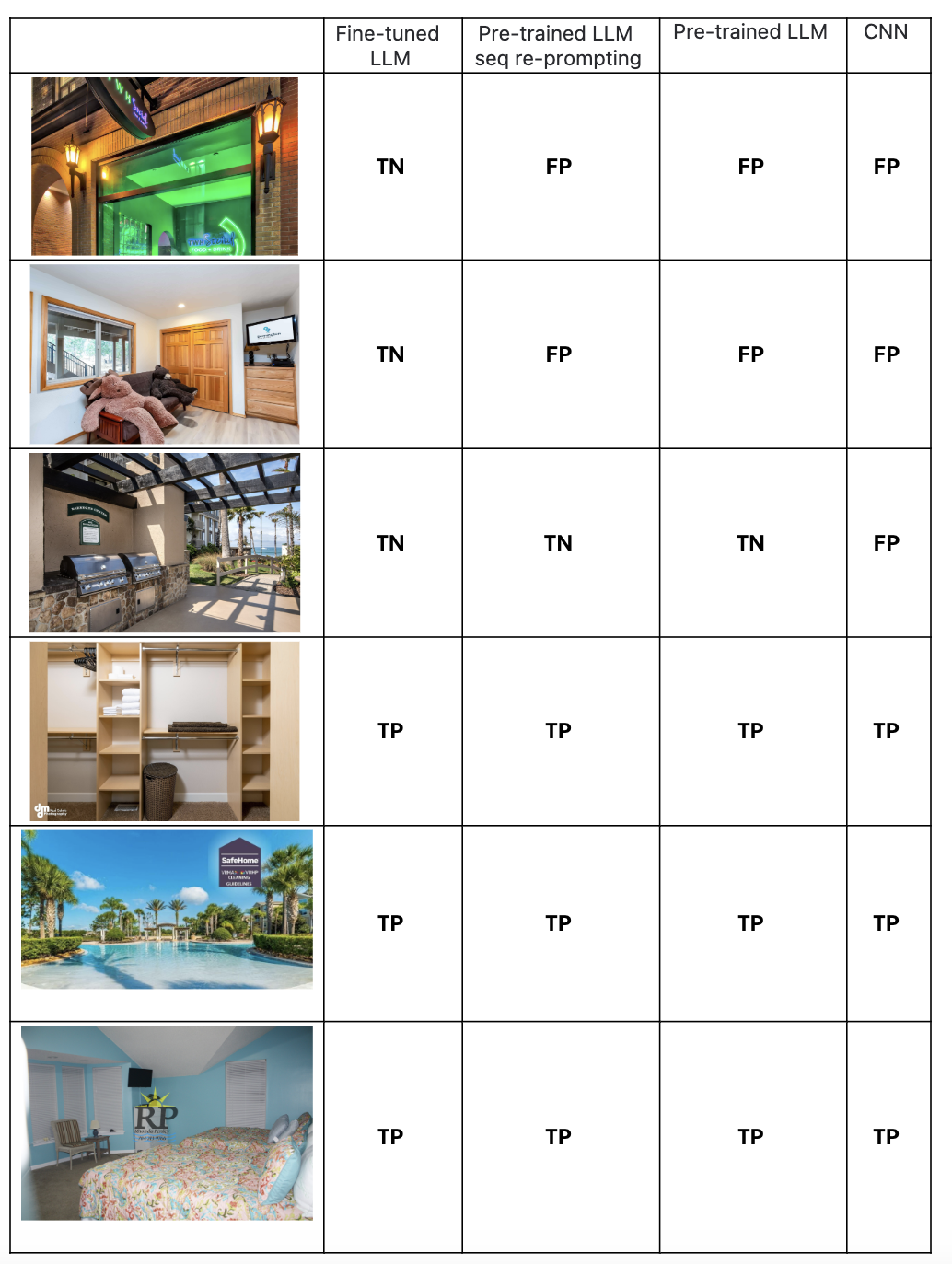}}
\caption{Performance of the models on some challenging images}
\label{fig:performance}
\end{figure}

Figure \ref{fig:performance} presents a selection of both challenging/non-challenging positive and negative samples, including images with overlaid text and natural signage. As illustrated, the fine-tuned LLM demonstrates an improved ability to distinguish between artificially imposed text overlays and embedded text elements such as signage. This indicates the model's enhanced contextual understanding and its capacity to differentiate between synthetic and naturally occurring textual features within the input images. 

\section{Conclusion}
Our study demonstrates that fine-tuning multi-modal LLMs on fewer than 1,000 images over just two epochs can significantly outperform both traditional CNNs and zero-shot LLM approaches in complex object detection tasks. The fine-tuned LLM achieved up to 36 percent higher accuracy compared to CNN baselines, along with notably higher precision—effectively reducing false positives and ensuring more reliable detection. These findings underscore the efficiency, adaptability, and data efficiency of language-guided models, particularly in low-resource settings. While our focus was on artificial text overlay detection, the proposed fine-tuning strategy is broadly applicable and can be extended to other complex vision-language tasks that require nuanced contextual understanding. Overall, this work presents a practical and scalable approach to bridging vision and language with minimal supervision.

\vspace{12pt}

\end{document}